\documentclass{bmvc2k}
\usepackage{algorithm}
\usepackage{algorithmic}
\usepackage{dsfont}
\usepackage{multirow} 
\usepackage{booktabs}
\usepackage{ragged2e}
\usepackage{amssymb}
\usepackage{pifont}
\usepackage{makecell}
\usepackage{tabularx}
\usepackage{geometry}
\usepackage{graphicx}
\usepackage{longtable}
\usepackage{tablefootnote}

\title{PSScreen: Partially Supervised Multiple Retinal Disease Screening}

\addauthor{Boyi Zheng}{boyi.zheng@oulu.fi}{}
\addauthor{Qing Liu$^{*}$}{qing.liu@oulu.fi}{}

\addinstitution{
 Center for Machine Vision and Signal
 Analysis (CMVS)\\
 University of Oulu\\
 Oulu, Finland
}

\runninghead{ZHENG, LIU}{PSScreen: Partially Supervised Retinal Disease Screening}

\def\eg{\emph{e.g}\bmvaOneDot}
\def\Eg{\emph{E.g}\bmvaOneDot}
\def\etal{\emph{et al}\bmvaOneDot}

\begin{document}

\maketitle
\footnotetext[1]{* Corresponding author}

\begin{abstract}

Leveraging multiple partially labeled datasets to train a model for multiple retinal disease screening reduces the reliance on fully annotated datasets, but remains challenging due to significant domain shifts across training datasets from various medical sites, and the label absent issue for partial classes. To solve these challenges, we propose \textbf{PSScreen}, a novel \textbf{P}artially \textbf{S}upervised multiple retinal disease \textbf{Screen}ing model. Our PSScreen consists of two streams and one learns deterministic features and the other learns probabilistic features via uncertainty injection. Then, we leverage the textual guidance to decouple two types of features into disease-wise features and align them via feature distillation to boost the domain generalization ability. Meanwhile, we employ pseudo label consistency between two streams to address the label absent issue and introduce a self-distillation to transfer task-relevant semantics about  known classes from the deterministic to the probabilistic stream to further enhance the detection performances. Experiments show that our PSScreen significantly enhances the detection performances on six retinal diseases and the normal state averagely and achieves state-of-the-art results on both in-domain and out-of-domain datasets. Codes are available at \url{https://github.com/boyiZheng99/PSScreen}.
\end{abstract}

\section{Introduction}

\begin{figure*}[!t] 
    \centering 
    \includegraphics[width=1\textwidth]{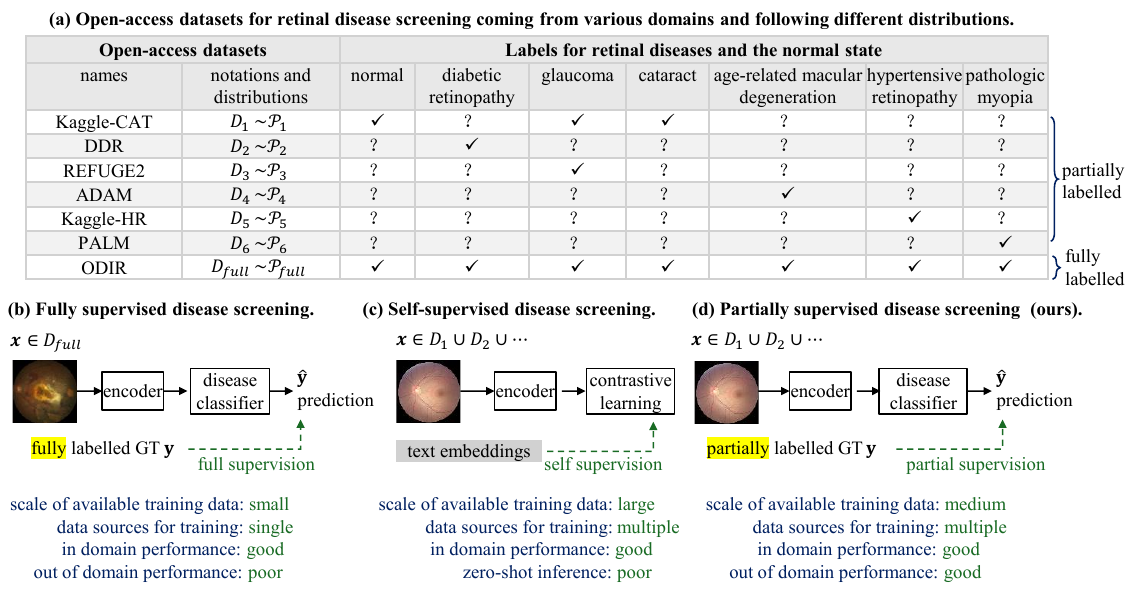} 
    \caption{Exampled open-access datasets for retinal disease screening and  screening model comparisons under three learning paradigms. (a) lists open-access datasets where "\checkmark" indicates labels for diseases are available while "?" denotes labels are not available. From (b) to (d), we illustrate the pipeline and characteristics of the fully supervised screening model usually trained with a fully labeled dataset, the self-supervised screening model trained with multiple datasets consisting of image-text pairs, and the partially supervised screening model trained with multiple partially labeled datasets. } 
    \label{fig:motivation} 
\end{figure*}

Automated detection of retinal disease with fundus images is crucial for efficient and cost-effective large-scale population screening. Fueled by the release of open-access datasets summarized in Fig.~\ref{fig:motivation}(a) for specific retinal diseases such as DDR \cite{Li2019} for diabetic retinopathy screening, REFUGE2 \cite{Fang2022a} for glaucoma screening, PALM \cite{fang2024open} for myopia screening etc., numerous works \cite {Sun2021,Das2023,Zhang2024a,Madarapu2024} have been developed to train disease-specific screening models on individual training datasets. Although these models are promising in screening for specific diseases on images within specific domains, they are still far from real-world applications where screening for as many retinal diseases as possible on images from various or even unseen domains is desired. Developing  a screening model for multiple retinal diseases  with strong domain generalization ability is of great significance, yet remains challenging.

The most intuitive way is to train a disease screening model in a fully supervised way with fully labeled training data as illustrated in Fig.~\ref{fig:motivation}(b). For example, TrustDetector \cite{TrustDetector} trains a screening model for three retinal diseases as the dataset collected only provides labels for three diseases while the method in \cite{OIA-ODIR-2021} trains the model on ODIR for multiple retinal disease screening. Although they achieve promising performances, the size of training data is limited and they assume that training and test images share the same distribution, which limits their generalization ability to out-of-domain data. The second way is to collect and manually annotate a large-scale dataset e.g. Retina-1M \cite{Ju2024}, to enable the fully supervised training. However, the annotation is labor-intensive and costly. More recently, with the adaptability of foundation models such as FLAIR \cite{SilvaRodriguez2025} and RET-CLIP \cite{Du2024} which are trained with large-scale image-text pairs in a self-supervised learning way as illustrated in Figure~\ref{fig:motivation}(c), zero-shot learning for retinal disease screening is emerging. Although large-scale fully annotated training data is not required, the performances for specific disease screening is poor \cite{MedIA-Zhang-2024}.

In this paper, we propose \textbf{PSScreen}, a \textbf{P}artially \textbf{S}upervised multiple retinal disease \textbf{Screen}ing model. As illustrated in Fig.~\ref{fig:motivation}(d), our PSScreen trains a screening model with multiple partially labeled datasets following different data distributions. Unlike previous partially supervised learning methods for natural images \cite{chen2022,pu2022semantic,Chen2024,Pu2024} and medical images \cite{Xiao2024}, where both training and testing data come from the same dataset without domain shift, the partially supervised learning in this paper is more challenge due to (1) domain shifts among training datasets collected from various medical sites, (2) the label absent issue for partial classes. To address them, we propose a two-stream network: one learns deterministic features and the other learns probabilistic features via uncertainty injection. Then, under the guidance of textual information, we decouple two types of features into disease-wise features and align them via feature distillation so that the features learned are robust against the domain shifts. Meanwhile, PSScreen addresses the label absent issue via pseudo label consistency between two streams. Finally, we introduce a self-distillation to transfer task-relevant semantics about known classes from the deterministic to the probabilistic stream to further enhance the detection performances. Experiments show that PSScreen outperforms state-of-the-art methods on in-domain and out-of-domain datasets.

To summarize, our contributions are as follows:
\begin{itemize}
    \item We propose PSScreen, a multiple retinal disease screening method which trains the disease screening model on a meta-dataset composed of multiple partially labeled datasets following different distributions. To the best of our knowledge, we are the first to train a partially supervised model for screening multiple retinal diseases using multiple datasets from various medical sites.
    \item We propose a two-stream network that learns deterministic features and uncertainty-injected probabilistic features and aligns them to boost the domain generalization ability at the feature level. Meanwhile, we introduce pseudo label consistency between two streams to address the label absent issue and a self-distillation to transfer the task-relevant semantics about known classes from the deterministic to the probabilistic stream to further enhance the detection performances.
    \item We validate our PSScreen on the meta-dataset, and demonstrate that our PSScreen achieves state-of-the-art performances. More importantly, validation on six unseen datasets further shows that PSScreen achieves superior domain generalization capability over previous methods.
\end{itemize}

\section{Proposed Method}

\textbf{Problem Formulation.} We suppose that (1) there is a meta-dataset $\mathcal{D}=\left\{D_1,D_2,\dots,D_K\right\}$ consisting of $K$ partially labeled datasets which are collected from various medical sites and follow different distributions, (2) each dataset $D_k = \{(\mathbf{x}_i, \mathbf{y}_i)\}_{i=1}^{N_k}$ with $N_k$ samples where $\mathbf{x}_i$ is the $i$-th sample and  $\mathbf{y}_i\in \{1, 0, -1\}^T$ is the label for $T$ retinal diseases and $y_{i,t}=0$ indicating that the $t$-th disease label is unknown, while $y_{i,t}=1/-1$ indicating the $t$-th disease label is positive/negative. For simplicity, we use the vector $\mathbf{\delta}_i=\mathds{1}_{\{1,-1\}}(\mathbf{y}_i)$ to indicate whether the label for diseases is known or unknown, where $\mathds{1}(\cdot)$ is an indicator function.  Our goal is to train a multi-label disease screening model on $\mathcal{D}$, which can predict risks for $T$ diseases and well generalize to out-of-domain test data. 



\begin{figure*}[!t] 
    \centering 
    \includegraphics[width=1\textwidth]{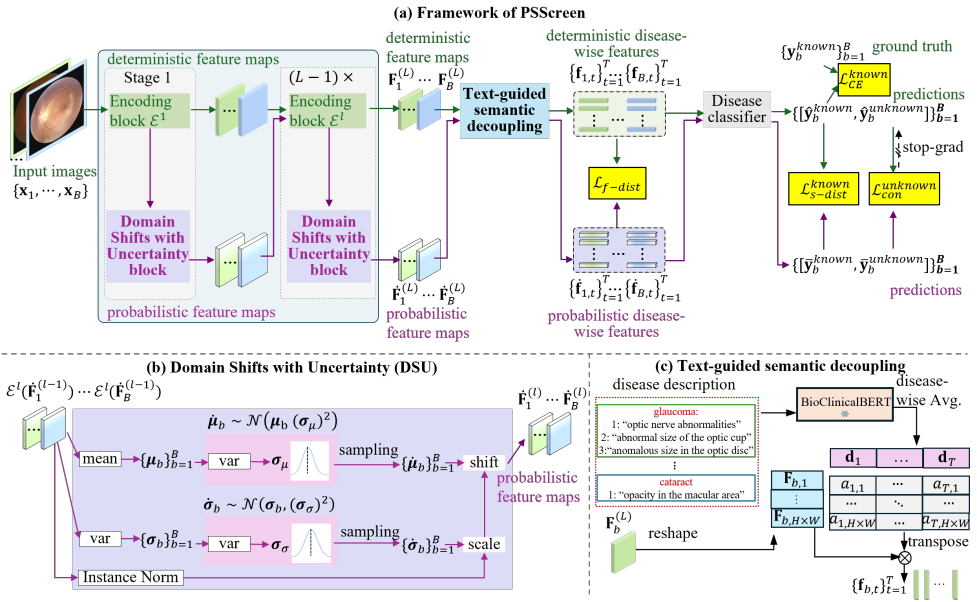} 
    \caption{The framework of \textit{PSScreen}. (a) illustrates the training pipeline of PSScreen. With training images, deterministic features and probabilistic features are extracted by the encoding blocks and domain shifts with uncertainty (DSU) blocks, then decoupled by the text-guided semantic decoupling module, finally fed to the disease classifier for multi-disease risk prediction. Feature distillation $\mathcal{L}_{f\text{-}dist}$, self-distillation for known classes $\mathcal{L}_{s\text{-}dist}^{known}$, pseudo label consistency for unknown classes $\mathcal{L}_{con}^{unknown}$, and cross entropy loss for known classes $\mathcal{L}_{CE}^{known}$ are applied for model optimization. (b) and (c) illustrate details for DSU and text-guided semantic decoupling block.} 
    \label{fig:framework} 
\end{figure*}

\textbf{Overview.} Fig.~\ref{fig:framework} illustrates the overview of \textit{PSScreen}. PSScreen is a two-stream network composed of six key modules: 1) Deterministic feature learning which outputs deterministic feature maps to preserve task-relevant semantics, 2) Probabilistic feature learning via attaching Domain Shifts with Uncertainty (DSU) block \cite{Li2022a} at each stage of the backbone to produce probabilistic feature maps, 3) Text-guided semantic decoupling which decouples global feature maps from the two streams into disease-wise features respectively, 4) Feature distillation which aligns deterministic and probabilistic disease-wise features in the latent space to boost the domain generalization ability, 5) Self-distillation which transfers task-relevant semantics about known classes from the deterministic to the probabilistic stream to enhance detection performances, and 6) Pseudo label consistency which addresses the label absent issue.




\textbf{Deterministic Feature Learning.}
We adopt either a $L$-stage CNN or vision transformer as the backbone to learn deterministic features from each batch of input images and obtain deterministic feature maps $\{\mathbf{F}_1^{(l)}, \cdots, \mathbf{F}_B^{(l)}\}$ where $l\in [1,\cdots, L]$. As deterministic feature maps preserve task-relevant semantics without uncertainty injection, the classification loss for known classes is imposed on predictions derived from them.

\textbf{Probabilistic Feature Learning via Domain Shifts with Uncertainty (DSU) Block.}
We estimate the uncertainty via attaching a DSU block \cite{Li2022a} into each stage of the backbone and produce probabilistic features to enhance the generalization ability across diverse or even unseen domains. As shown in Fig.~\ref{fig:framework}(b), probabilistic feature maps $\{\mathbf{\dot{F}}_1^{(l-1)}, \cdots, \mathbf{\dot{F}}_B^{(l-1)}\}$ from the $(l{-}1)$-th stage are first fed into the $l$-th encoding block, yielding ${\varepsilon^l(\mathbf{\dot{F}}_1^{(l-1)}), \cdots, \varepsilon^l(\mathbf{\dot{F}}_B^{(l-1)})}$. Then, we calculate channel-wise means $\{\mathbf{\mu}_b\}_{b=1}^B$ and variances $\{\mathbf{\sigma}_b\}_{b=1}^B$, and model $\mathbf{\mu}_b$ and $\mathbf{\sigma}_b$ as centers of two independent Gaussian distributions. Their respective scopes, i.e., the uncertainty for the mean $\mathbf{\sigma}_{\mu}$ and uncertainty for the variance $\mathbf{\sigma}_{\sigma}$, are estimated non-parametrically which are variances of $\{\mathbf{\mu}_b\}_{b=1}^B$ and $\{\mathbf{\sigma}_b\}_{b=1}^B$ respectively. Finally, we randomly sample the mean and variance from these distributions and scale the instance normalized features with the sampled variance and shift them with the sampled mean to obtain the probabilistic feature maps $\{{\mathbf{\dot{F}}_1^{(l)}, \cdots, \mathbf{\dot{F}}_B^{(l)}}\}$ where $l\in [1,\cdots, L]$.

\textbf{Text-guided Semantic Decoupling.} To focus on disease-specific semantic regions, we introduce a text-guided semantic decoupling module inspired by \cite{Chen2019a}. As illustrated in Fig~\ref{fig:framework}(c), we use BioClinicalBERT\footnote{\url{https://huggingface.co/emilyalsentzer/Bio_ClinicalBERT}} to encode multiple expert knowledge descriptions of each disease from \cite{SilvaRodriguez2025} and average them to obtain the disease-wise text embeddings $\{\mathbf{d}_t\}_{t=1}^T$. For the visual features $\mathbf{F}_b^{(L)}$ with spatial size $H\times W$, we first reshape them into $ \{\mathbf{F}_{b,i}\}_{i=1}^{H\times W}$, then calculate the attention score $\alpha_{t, i}$ between each visual feature $\mathbf{F}_{b,i}$ and text feature $\mathbf{d}_t$ via:
\begin{equation}
\alpha_{t, i}  = \frac{\text{exp} \left( \mathbf{v}_{att}^\top \tanh\left( \mathbf{W}_{att}^F \mathbf{F}_{b,i} \odot \mathbf{W}_{att}^d \mathbf{d}_t  \right)\right)}{\sum_{j=1}^{H\times W} \text{exp} \left(  \left( \mathbf{v}_{att}^\top \tanh\left( \mathbf{W}_{att}^F \mathbf{F}_{b,j} \odot \mathbf{W}_{att}^d \mathbf{d}_t \right) \right)\right)} \;,
\end{equation}
where $\odot$ denotes the Hadamard product, and $\mathbf{W}_{att}^F$, $\mathbf{W}_{att}^d$, and $\mathbf{v}_{att}$ are learnable weights. Finally, we obtain the text-guided disease-wise features via:
\begin{equation}
\mathbf{f}_{b,t} =\sum_{i=1}^{H\times W} \alpha_{t, i} \cdot \mathbf{F}_{b,i}\;.
\end{equation}

\textbf{Feature Distillation.} To align feature distributions, we minimize the maximum mean discrepancies (MMD) loss \cite{Long2015} which measures the discrepancy between two distributions by comparing their means in a reproducing kernel Hilbert space $\mathcal{H}$. Specifically, we minimize the MMD loss between the text-guided semantic  decoupled features from the two-streams: 
\begin{equation}
\mathcal{L}_{f\text{-}dist}(\mathbf{f}_{b,t}, \mathbf{\dot{f}}_{b,t}) =  \frac{1}{T} \sum_{t=1}^T \left\lVert\phi(\mathbf{f}_{b,t}) 
- \phi(\mathbf{\dot{f}}_{b,t}) 
\right\rVert_{\mathcal{H}}^2,
\label{MMD_loss}
\end{equation}
where $\phi(\cdot)$ denotes the kernel mapping function. In practice, we use a Gaussian kernel to compute the MMD loss. This class-specific alignment ensures that probabilistic disease-wise features closely align with their deterministic counterparts while preserving essential class-specific discriminative information. By aligning the two-stream features, the learned features help the model boost the domain generalization ability.

\textbf{Classification Loss for Known Classes.} We follow \cite{Durand2019} and use the partial binary cross entropy loss as the classification loss for known classes: 
\begin{align}
\mathcal{L}_{CE}^{known}(\mathbf{y}_b^{known},\hat{\mathbf{y}}_b^{known}) 
= - \frac{1}{\|\delta_b\|_1} 
\sum_{t=1}^T \left( \mathds{1}_{\{1\}}(y_{b,t}) \log(\hat{y}_{b,t}) 
+ \mathds{1}_{\{-1\}}(y_{b,t}) \log(1 - \hat{y}_{b,t}) \right),
\end{align}
where $\hat{y}_{b,t}$ is the prediction corresponding to the deterministic disease-wise feature for $t$-th disease of the $b$-th image in current batch, and  $y_{b,t}$ is the ground-truth label.

\textbf{Self-distillation.} To supervise the probabilistic feature learning, we introduce self-distillation to transfer the task-relevant semantics from the deterministic to the probabilistic stream via aligning the two-stream network’s output distributions of the known classes. To this end, we minimize the KL divergence loss between the final classification predictions of the two streams for known classes:
\begin{equation}
    \mathcal{L}_{s\text{-}dist}^{known}\left(\hat{\mathbf{y}}_b^{known},\bar{\mathbf{y}}_b^{known}\right) = KL\left(\hat{\mathbf{y}}_b^{known}|| \bar{\mathbf{y}}_b^{known}\right) =  -\frac{1}{||\mathbf{\delta}_b||_1}\sum_{t=1}^{T}\mathbf{\delta}_{b,t}\cdot 
    \hat{y}_{b,t} \cdot \log \frac{\bar{y}_{b,t}}{\hat{y}_{b,t}},
\label{sd_label_loss}
\end{equation}
where $\bar{y}_{b,t}$ is the prediction corresponding to the probabilistic disease-wise feature for $t$-th disease of the $b$-th image in current batch.

\textbf{Pseudo Label Consistency.} We further enforce the consistency between the pseudo labels derived from the probabilistic features for unknown classes and those derived from deterministic features. In detail, we adopt confidence-based “hard” pseudo labels and samples with predictions greater than a threshold $\tau$ are treated as positive and those below $1 - \tau$ are treated as negative. With them, the pseudo label consistency loss can be expressed as:
\begin{align}
\mathcal{L}_{con}^{unknown}(\hat{\mathbf{y}}_b^{unknown},\bar{\mathbf{y}}_b^{unknown}) & \notag\\ = -\frac{1}{T-||\mathbf{\delta}_b||_1}\sum_{t=1}^T &\left(1-\mathbf{\delta}_{b,t}\right)\cdot \left(\mathds{1}_{(\hat{y}_{b,t} > \tau)}\log(\bar{y}_{b,t})+ \mathds{1}_{(\hat{y}_{b,t} < 1-\tau)} \log(1 - \bar{y}_{b,t})\right) \;.
\label{sd_unlabel_loss}
\end{align}
To ensure that the generated pseudo labels for unknown classes are sufficiently accurate, $\tau$ is set as 0.95 in our experiment.

\textbf{Total Loss.} The total loss function is :
\begin{equation}
\mathcal{L} = \mathcal{L}_{CE}^{known} + \lambda_1 \mathcal{L}_{f\text{-}dist} + \lambda_2 \mathcal{L}_{s\text{-}dist}^{known} + \lambda_3 \mathcal{L}_{con}^{unknown}, 
\label{overall_loss}
\end{equation}
where $\lambda_1$ and $\lambda_2$ are set to 0.05 and 1 respectively, in order to ensure that the different loss components are on a comparable scale. $\lambda_3$ is used to control the contribution of the pseudo label loss during training. It is set to 0 for the first 5 epochs and updated to 0.6 starting from the 6-th epoch.

\section{Experiment}
\subsection{Experimental Settings}
\textbf{Datasets.} We construct two combined datasets using multiple partially labeled open-access datasets: (1)  \textbf{meta-dataset}, constituting of six datasets, i.e., DDR \cite{Li2019}, ADAM \cite{Fang2022}, PALM \cite{fang2024open}, Kaggle-CAT\footnote{\url{https://www.kaggle.com/datasets/jr2ngb/cataractdataset}\label{cat}}, Kaggle-HR\footnote{\url{https://www.kaggle.com/datasets/harshwardhanfartale/hypertension-and-hypertensive-retinopathy-dataset}\label{HR}} and REFUGE2 \cite{Fang2022a}, and (2) \textbf{unseen-dataset} constituting of four datasets i.e., APTOS2019 \footnote{\url{https://www.kaggle.com/competitions/aptos2019-blindness-detection/data}\label{aptos}}, ORIGA$^{\text{light}}$ \cite{Zhang2010}, HPMI \cite{Huang2023}, and RFMiD \cite{Pachade2021}.
The meta-dataset covers all diseases of interest, including diabetic retinopathy (DR), glaucoma, cataract, age-related macular degeneration (AMD), hypertensive retinopathy (HR), pathologic myopia (PM), and the normal state. It is utilized for both training and in-domain validation. The unseen-dataset is used for out-of-domain validation. Besides, the test set of \textbf{ODIR dataset} \cite{OIA-ODIR-2021} is also used for out-of-domain validation. To further evaluate the model’s domain generalization ability, we follow \cite{SilvaRodriguez2025} and use \textbf{ODIR200$\times$3} to validate the performances under the setting of zero-shot inference. ODIR200$\times$3 is a 600-image subset containing three diseases, i.e., the normal state, cataract, and pathologic myopia and with 200 images in each category. More details about the meta-dataset, unseen-dataset and ODIR200x3 can be found in Supplementary \ref{sec:sppl_dataset}.





\textbf{Evaluation Metrics.} F-score and quadratic weighted kappa (QWK) are adopted following \cite{OIA-ODIR-2021,Zhang2024a}. For evaluation across multiple tasks and datasets, we compute the average F-score as:
\begin{equation}
mF= \sum_{t=1}^{T} \sum_{k=1}^{K^{(t)}} \frac{1}{T} \frac{1}{K^{(t)}} F^{(t)}_{k},
\end{equation}
where $T$ is the number of tasks and $K^{(t)}$ is the number of datasets per task. Mean QWK ($mQWK$) is calculated similarly.

\textbf{Implementation Details.} We crop the field of view from each fundus image, then pad the short side with zeros to equal length with the long side and resize it to 512×512. For augmentation, we apply random scaling with a scale factor uniformly sampled from [0.8,1.2] with a probability of 0.5, followed by padding or cropping to maintain the input size. We then apply the augmentation strategies from \cite{Rodriguez2023}, excluding Cutout. Following \cite{chen2022,pu2022semantic,Chen2024,Pu2024}, we adopt ResNet-101 \cite{He2016} pretrained on ImageNet \cite{Deng2009} as the backbone, with other model parameters initialized randomly. Training uses the ADAM optimizer \cite{Kingma2015} with a batch size of 16, weight decay of 5×10$^{\text{-4}}$, and an initial learning rate of 1×10$^{\text{-5}}$ reduced by a factor of 10 every 10 epochs. PSScreen is trained with 20 epochs in total, and is implemented with PyTorch on one NVIDIA A100 GPU with 40 GB RAM.

\begin{table}[!t]
\centering
\setlength{\tabcolsep}{4pt} 
\resizebox{0.95\textwidth}{!}{
\begin{tabular}{|l|c|cc|cc|cc|c|c|c|}
\hline
\multirow{2}{*}{Methods} & \multirow{2}{*}{Source} & \multicolumn{2}{c|}{Meta} & \multicolumn{2}{c|}{Unseen} & \multicolumn{2}{c|}{ODIR} & FPS & GFLOPs & \#param (M) \\
\cline{3-8}
& & $mF$ & $mQWK$ & $mF$ & $mQWK$ & $mF$ & $mQWK$ & & & \\
\hline
Full supervise & -- & -- & -- & -- & -- & 69.8\scriptsize$\pm$0.2 & 52.6\scriptsize$\pm$0.7 & -- & -- & -- \\
MultiNets & -- & 82.7\scriptsize$\pm$0.7 & 73.0\scriptsize$\pm$1.4 & 62.4\scriptsize$\pm$1.2 & 44.8\scriptsize$\pm$2.3 & 56.4\scriptsize$\pm$1.4 & 26.1\scriptsize$\pm$2.1 & 47.7 & 5240 & 837.4 \\
MultiHeads & -- & 83.6\scriptsize$\pm$0.7 & 75.0\scriptsize$\pm$1.3 & 62.4\scriptsize$\pm$1.5 & 44.2\scriptsize$\pm$2.0 & 56.7\scriptsize$\pm$0.6 & 27.7\scriptsize$\pm$0.4 & 382.6 & 650 & 105.3 \\
SST \cite{chen2022} & AAAI22 & 83.4\scriptsize$\pm$1.4 & 74.7\scriptsize$\pm$2.8 & \underline{63.2\scriptsize$\pm$0.9} & \underline{46.7\scriptsize$\pm$1.8} & 55.6\scriptsize$\pm$1.7 & 26.2\scriptsize$\pm$2.5 & 357.9 & 670 & 122.9 \\
SARB \cite{pu2022semantic} & AAAI22 & \underline{84.0\scriptsize$\pm$0.9} & \underline{75.7\scriptsize$\pm$1.8} & 60.5\scriptsize$\pm$1.0 & 40.3\scriptsize$\pm$1.2 & \underline{61.4\scriptsize$\pm$0.9} & \underline{33.8\scriptsize$\pm$1.6} & 361.6 & 670 & 112.7 \\
BoostLU \cite{kim2023} & CVPR23 & 80.3\scriptsize$\pm$0.5 & 69.5\scriptsize$\pm$1.1 & 44.5\scriptsize$\pm$0.8 & 18.0\scriptsize$\pm$0.6 & 50.6\scriptsize$\pm$1.0 & 17.5\scriptsize$\pm$2.1 & 140.5 & 650 & 162.2 \\
HST \cite{Chen2024} & IJCV24 & 83.4\scriptsize$\pm$0.4 & 74.7\scriptsize$\pm$0.8 & 62.3\scriptsize$\pm$1.7 & 44.8\scriptsize$\pm$3.8 & 56.0\scriptsize$\pm$0.8 & 27.5\scriptsize$\pm$0.8 & 357.9 & 670 & 123.0 \\
CALDNR \cite{Pu2024} & TMM24 & 82.0\scriptsize$\pm$1.1 & 72.8\scriptsize$\pm$2.2 & 47.6\scriptsize$\pm$3.1 & 19.0\scriptsize$\pm$6.7 & 52.1\scriptsize$\pm$0.6 & 19.1\scriptsize$\pm$0.8 & 138.4 & 660 & 120.9 \\
\textbf{PSScreen (ours)} & -- & \textbf{84.2\scriptsize$\pm$0.3} & \textbf{76.8\scriptsize$\pm$0.8} & \textbf{65.9\scriptsize$\pm$0.1} & \textbf{50.9\scriptsize$\pm$0.1} & \textbf{64.1\scriptsize$\pm$1.0} & \textbf{39.8\scriptsize$\pm$1.3} & 373.1 & 660 & 116.8 \\
\hline
\end{tabular}
}
\caption{Performance comparison of different partially supervised learning methods on three datasets. The best and second-best are highlighted in bold and with an underline. Means and standard deviations are reported over three trials. Results for each disease can be found in Supplementary \ref{sppl:class-wise-rst}.}
\label{tab:performance_PSL}
\end{table}

\subsection{Results}

\textbf{Comparison with Partially Supervised Learning Methods.} We compare our PSScreen with two baseline approaches, MultiNets and MultiHeads, as well as five state-of-the-art (SOTA) methods: SST \cite{chen2022}, SARB \cite{pu2022semantic}, HST \cite{Chen2024}, BoostLU \cite{kim2023}, and CALDNR \cite{Pu2024}. MultiNets trains multiple task-specific models for each task and combines all prediction results during testing. MultiHeads consists of a backbone network and a classifier, which outputs prediction probabilities for all classes.

We report performances in Table~\ref{tab:performance_PSL}, and observe that on the meta-dataset (1) our PSScreen achieves the best, (2) our PSScreen outperforms the second-best SARB \cite{pu2022semantic} by 0.2\% in $mF$  and 1.1\% in $mQWK$. Notably, results on two out-of-domain datasets, i.e., unseen and ODIR \cite{OIA-ODIR-2021} show that PSScreen gains significant improvements compared to other methods. In detail, on the unseen-dataset, our PSScreen outperforms the second-best SST \cite{chen2022} by 2.7\% in $mF$ and 4.2\% in $mQWK$. On the ODIR \cite{OIA-ODIR-2021}, PSScreen outperforms the second-best SARB \cite{pu2022semantic} by 2.7\% in $mF$ and 6.0\% in $mQWK$. For reference, we also report the performances on the ODIR \cite{OIA-ODIR-2021} by the fully supervised model as an upper bound. SOTAs in Table~\ref{tab:performance_PSL} perform poorly on the two out-of-domain datasets as they assume that training and test images follow the same distribution. Additionally, we compare model efficiency and report Frames Per Second (FPS), Giga Floating Point Operations (GFLOPs), and number of parameters (\#param). It can be observed that the FPS of PSScreen is only slightly lower than that of MultiHeads, indicating comparable inference efficiency. Meanwhile, compared to SOTAs, PSScreen significantly improves prediction accuracy without introducing additional computational or parameter overhead.


\begin{figure}
    \centering
    \includegraphics[width=1.0\linewidth]{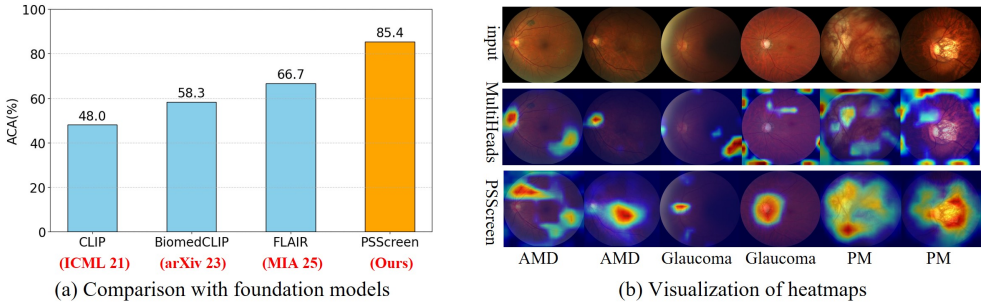}
    \caption{(a) Performance comparison of zero-shot inference with foundation models on the ODIR200x3 dataset. (b) Visualization of heatmaps generated by MultiHeads and PSScreen for three retinal diseases: age-related macular degeneration (AMD), glaucoma, and pathologic myopia (PM).}
    \label{fig:zero-s_visual}
\end{figure}

\textbf{Comparison of Zero-shot Inference with Foundation Models.} We further compare the zero-shot performances of our PSScreen and three latest foundation models, i.e., CLIP \cite{Radford2021}, BiomedCLIP \cite{Zhang2023}, and FLAIR \cite{SilvaRodriguez2025} on ODIR200$\times$3. Following FLAIR, we adopt ACA \cite{Zhao2019} as the evaluation metric and report ACAs in Fig.~\ref{fig:zero-s_visual}(a). As shown, foundation models perform suboptimally in zero-shot retinal disease screening. CLIP \cite{Radford2021} and BiomedCLIP \cite{Zhang2023} struggle to detect retinal diseases, while FLAIR \cite{SilvaRodriguez2025}  fails to generalize well to unseen diseases due to the complex pathological structures of retinal diseases. Our PSScreen achieves the best performance on ODIR200$\times$3, outperforming FLAIR by 18.7\%.


\textbf{Visualization.} We use GradCAM \cite{Selvaraju2017} to obtain heatmaps from the last convolutional layer of MultiHeads and PSScreen. As shown in Fig.~\ref{fig:zero-s_visual}(b), MultiHeads often fails to localize lesions, instead attending to domain-specific background. In contrast, our PSScreen consistently highlights lesion areas. For example, as shown in the first column, PSScreen accurately localizes AMD-related hemorrhage; in the second column, it identifies exudates critical for AMD diagnosis. For glaucoma, PSScreen consistently attends to the optic disc and cup, and for PM, PSScreen localizes retinal atrophy regions strongly correlated with the disease.

\subsection{Ablation Study}
\textbf{How $\mathcal{L}_{f\text{-}dist}$, $\mathcal{L}_{s\text{-}dist}^{known}$ and $\mathcal{L}_{con}^{unknown}$ Contribute?} There are three key loss terms, i.e., $\mathcal{L}_{f\text{-}dist}$ in Eq. \ref{MMD_loss}, $\mathcal{L}_{s\text{-}dist}^{known}$ in Eq. \ref{sd_label_loss} and $\mathcal{L}_{con}^{unknown}$ in Eq. \ref{sd_unlabel_loss}. To validate their effectiveness, we conduct ablation experiments and report the performance in Table~\ref{tab:core_module}. Removing the three loss items, PSScreen degrades into MultiHeads equiped with the text-guided semantic decoupling, which achieves inferior performance to PSScreen obviously. With $\mathcal{L}_{f\text{-}dist}$ and $\mathcal{L}_{s\text{-}dist}^{known}$ separately, in-domain performances and domain generalization are improved. On the contrary, $\mathcal{L}_{con}^{unknown}$ alone leads to the performance degradation which is possibly caused by  the loss of task-relevant semantics in probabilistic feature maps. Combining $\mathcal{L}_{s\text{-}dist}^{known}$ and $\mathcal{L}_{con}^{unknown}$ improves the model's performance. Integrating all loss terms achieves the best results, with $mQWK$ improvements of 5.5\% and 11.8\% on the unseen and ODIR datasets, respectively, compared to the degraded variant without the three loss terms, validating the effectiveness of PSScreen.


\textbf{Influences of Loss Weights $\lambda_1$, $\lambda_2$, and $\lambda_3$.} We vary $\lambda_1$, $\lambda_2$, and $\lambda_3$, and report $mQWK$ in Table~\ref{tab:backbone_mu}(a), which shows that $\lambda_1=0.05$, $\lambda_2=1.0$, and $\lambda_3=0.6$ perform the best performances. The performances on $mF$ can be found in Supplementary \ref{sppl:ablation_loss}.

\textbf{Compatibility with Different Backbones.} To further validate the compatibility of our PSScreen, we report $mQWK$ of PSScreen with different backbones including ConvNeXt-T \cite{Liu2022}, ConvNeXt V2-T \cite{Woo2023}, Swin-T \cite{Liu2021}, and VMamba-T \cite{Liu2024d} and compare them with the naive ones, i.e., MultiHeads with various backbones in Table~\ref{tab:backbone_mu}(b). The results demonstrate that our PSScreen consistently improves the performances across all three datasets with different backbones. The performances on $mF$ can be found in Supplementary \ref{sppl:ablation_backbone}.


\begin{table}[t]
\centering
 \resizebox{0.8\columnwidth}{!}{ 
 \begin{tabular}{|ccc|cc|cc|cc|}
        \hline
         $\mathcal{L}_{f\text{-}dist}$ &$\mathcal{L}_{s\text{-}dist}^{known}$ & $\mathcal{L}_{con}^{unknown}$ & \multicolumn{2}{c|}{Meta} & \multicolumn{2}{c|}{Unseen} & \multicolumn{2}{c|}{ODIR} \\
        \cline{4-5} \cline{6-7} \cline{8-9}
         & & & $mF$ & $mQWK$ & $mF$ & $mQWK$ & $mF$ & $mQWK$ \\
        \hline
        & & & 83.5\scriptsize$\pm$0.2 & 75.0\scriptsize$\pm$0.4 & 63.0\scriptsize$\pm$0.8 & 45.4\scriptsize$\pm$1.3 & 57.0\scriptsize$\pm$0.6 & 28.0\scriptsize$\pm$1.0 \\
        \checkmark & &  & 83.4\scriptsize$\pm$0.7 & 75.3\scriptsize$\pm$0.2 & 63.4\scriptsize$\pm$0.1 & 46.7\scriptsize$\pm$0.9 & 58.0\scriptsize$\pm$0.4 & 29.6\scriptsize$\pm$0.5 \\
         & \checkmark&  & 83.4\scriptsize$\pm$0.2 & 75.8\scriptsize$\pm$0.4 & 64.0\scriptsize$\pm$0.3 & 47.5\scriptsize$\pm$0.7 & 57.5\scriptsize$\pm$1.0 & 28.7\scriptsize$\pm$1.5 \\
        &  & \checkmark & 83.3\scriptsize$\pm$0.7 & 75.5\scriptsize$\pm$1.1 & 61.8\scriptsize$\pm$1.1 & 44.1\scriptsize$\pm$2.1 & 61.7\scriptsize$\pm$0.9 & 34.8\scriptsize$\pm$1.6 \\
         & \checkmark &\checkmark & 83.8\scriptsize$\pm$0.2 & 76.4\scriptsize$\pm$0.6 & 65.2\scriptsize$\pm$0.3 & 49.5\scriptsize$\pm$0.7 & 63.5\scriptsize$\pm$0.8 & 38.0\scriptsize$\pm$1.5 \\
        \checkmark & \checkmark & \checkmark & \textbf{84.2\scriptsize$\pm$0.3} & \textbf{76.8\scriptsize$\pm$0.8} & \textbf{65.9\scriptsize$\pm$0.1} & \textbf{50.9\scriptsize$\pm$0.1} & \textbf{64.1\scriptsize$\pm$1.0} & \textbf{39.8\scriptsize$\pm$1.3} \\
        \hline
    \end{tabular}}
\caption{The ablation study on the key loss terms of PSScreen.}
\label{tab:core_module}
\end{table}

\begin{table}[t!]
\centering
\begin{minipage}[t]{0.49\textwidth}
    \renewcommand{\arraystretch}{1.03} 
    \centering
    \resizebox{\textwidth}{!}{
    \begin{tabular}{|c|c|c|c|c|c|c|}
        \hline
        $\lambda_1$ & $\lambda_2$ & $\lambda_3$ & Meta & Unseen & ODIR \\
        \hline
        0.1   & 1.0   & 0.6 & 76.1\scriptsize$\pm$0.6 & 50.1\scriptsize$\pm$0.2 & 39.4\scriptsize$\pm$0.7 \\
        \textbf{0.05}  & \textbf{1.0}   & \textbf{0.6} & \textbf{76.8\scriptsize$\pm$0.8} & \textbf{50.9\scriptsize$\pm$0.1} & \textbf{39.8\scriptsize$\pm$1.3} \\
        0.025 & 1.0   & 0.6 & 76.7\scriptsize$\pm$0.1 & 49.4\scriptsize$\pm$0.6 & 39.7\scriptsize$\pm$1.0 \\
        \hline
        0.05  & 0.5   & 0.6 & 76.2\scriptsize$\pm$0.3 & 50.0\scriptsize$\pm$0.6 & 38.6\scriptsize$\pm$0.9 \\
        \textbf{0.05}  & \textbf{1.0}   & \textbf{0.6} & \textbf{76.8\scriptsize$\pm$0.8} & \textbf{50.9\scriptsize$\pm$0.1} & \textbf{39.8\scriptsize$\pm$1.3} \\
        0.05  & 2.0   & 0.6 & 75.9\scriptsize$\pm$0.4 & 48.2\scriptsize$\pm$1.3 & 38.7\scriptsize$\pm$0.5 \\
        \hline
        0.05  & 1.0   & 0.4 & 76.5\scriptsize$\pm$1.3 & 48.7\scriptsize$\pm$1.2 & 38.9\scriptsize$\pm$1.0 \\
        \textbf{0.05}  & \textbf{1.0}   & \textbf{0.6} & \textbf{76.8\scriptsize$\pm$0.8} & \textbf{50.9\scriptsize$\pm$0.1} & \textbf{39.8\scriptsize$\pm$1.3} \\
        0.05  & 1.0   & 0.8 & 76.6\scriptsize$\pm$0.5 & 50.5\scriptsize$\pm$1.4 & 38.8\scriptsize$\pm$0.4 \\
        \hline
        \end{tabular}
    }
    
    {\vspace{2pt} \small (a) $mQWK$ under different loss weight settings.}

\end{minipage}
\begin{minipage}[t]{0.48\textwidth}
    \centering
    \resizebox{\textwidth}{!}{
     \begin{tabular}{|l|rrr|}
        \hline
        Methods & Meta & Unseen & ODIR
        \\ \hline
        ResNet-101 \cite{He2016}        &     75.0\scriptsize$\pm$1.3               &      44.2\scriptsize$\pm$2.0               &  27.7\scriptsize$\pm$0.4     \\             
        +PSScreen              &       \textbf{76.8\scriptsize$\pm$0.8}              &      \textbf{50.9\scriptsize$\pm$0.1}                   &       \textbf{39.8\scriptsize$\pm$1.3}                \\ \hline
        ConvNeXt-T \cite{Liu2022}      &        75.0\scriptsize$\pm$0.7             &        48.2\scriptsize$\pm$1.2              &    25.4\scriptsize$\pm$2.3   \\               
        +PSScreen                &         \textbf{76.6\scriptsize$\pm$0.2}            &      \textbf{49.6\scriptsize$\pm$2.1}                     &              \textbf{37.0\scriptsize$\pm$1.5}       \\ \hline
        ConvNeXt V2-T \cite{Woo2023}          &      76.3\scriptsize$\pm$0.7               &         48.6\scriptsize$\pm$1.2                   &   28.9\scriptsize$\pm$1.3     \\              
        +PSScreen                   &     \textbf{77.0\scriptsize$\pm$0.2}                &       \textbf{49.1\scriptsize$\pm$0.5}               &        \textbf{38.3\scriptsize$\pm$1.0}               \\ \hline
        Swin-T \cite{Liu2021}          &      75.8\scriptsize$\pm$0.8               &         49.5\scriptsize$\pm$0.7                   &   25.4\scriptsize$\pm$0.3     \\              
        +PSScreen                   &     \textbf{79.3\scriptsize$\pm$0.7}                &       \textbf{50.8\scriptsize$\pm$0.6}               &        \textbf{35.5\scriptsize$\pm$0.9}               \\ \hline
        
        VMamba-T  \cite{Liu2024d}           &      76.3\scriptsize$\pm$1.0               &         47.2\scriptsize$\pm$0.5                   &   27.0\scriptsize$\pm$0.4     \\              
        +PSScreen                   &     \textbf{78.0\scriptsize$\pm$0.7}                &       \textbf{49.8\scriptsize$\pm$1.7}               &        \textbf{36.9\scriptsize$\pm$1.8}               \\
        \hline
    \end{tabular}
    }
    {\vspace{2.5pt} \small (b) Performances with different backbones.}
\end{minipage}
\caption{(a) $mQWK$ under different settings of $\lambda_1$, $\lambda_2$, and $\lambda_3$. (b) $mQWK$ by  MultiHeads and PSScreen with different backbones. Results in terms of $mF$ can be found in Supplementary \ref{sppl:ablation_loss}.}
\label{tab:backbone_mu}
\end{table}

\section{Conclusion and Future Work}

In this paper, we propose \textit{PSScreen}, a multiple retinal disease screening model trained on multiple partially labeled datasets following different distributions. PSScreen is a two-stream network and one streams learns deterministic features and the other learns probabilistic features via DSU blocks. The features are then decoupled by the text-guided semantic decoupling modules to facilitate multiple disease detection. To enforce the probabilistic feature learning stream to learn task-relevant semantics, feature distillation and self-distillation are employed to transfer task-relevant semantics from the deterministic to the probabilistic stream. Meanwhile, pseudo label consistency is imposed to address label absent issue. Extensive experiments on both fully and partially labeled datasets demonstrate that PSScreen achieves state-of-the-art performance across multiple in-domain and out-of-domain datasets.

PSScreen has a lightweight architecture and fast inference speed, making it well-suited for integration into clinical workflows. However, challenges remain in translating screening models from bench to bedside. For example, ophthalmic imaging modalities are diverse, whereas our current focus is limited to fundus images. To enhance clinical applicability, incorporating additional modalities for screening a broader range of retinal diseases is desirable in the future. Even though  GradCAM-based visualizations for model's interpretability are provided, a structured evaluation involving clinicians is lacking. Moreover, how to conduct clinician-in-the-loop assessments to verify the model's interpretability remains still an open question and merits further investigation.

\section{Acknowledgments}
This work was supported in part by Academy Research Fellow project under Grant 355095 and Hunan Provincial Natural Science Foundation of China under Grant 2023JJ30699. As well, the authors wish to acknowledge the CSC–IT Center for Science, Finland, for computational resources.

\bibliography{BMVC}
\newpage


\setcounter{table}{3}
\setcounter{figure}{3}

\section*{Supplementary for PSScreen: Partially Supervised Multiple Retinal Disease Screening}
\label{sppl}


\setcounter{section}{0}    
\renewcommand{\thesection}{\Alph{section}}  

\def\eg{\emph{e.g}\bmvaOneDot}
\def\Eg{\emph{E.g}\bmvaOneDot}
\def\etal{\emph{et al}\bmvaOneDot}

\section{Details of Open-access Datasets.}
\label{sec:sppl_dataset}
Table \ref{table1:dataset_info} presents the details of all the open-access datasets used in our study. To establish a protocol for the experimental evaluation, we process a subset of them. RFMiD \cite{Pachade2021} contains 46 retinal disease classes or structural abnormalities of which only the normal state and AMD classes are used in our study. For ODIR \cite{OIA-ODIR-2021}, "other diseases" class is excluded as it is irrelevant to our study and problematic images (such as those affected by lens dust or poor quality) are exluceded following the ODIR official supplementary \footnotemark[1], yielding a dataset of 6,961 training images and 988 testing images. 

\begin{table*}[!h]
    \centering
    \Large
    \resizebox{1.0\textwidth}{!}{
    \begin{tabular}{|c|l|p{7cm}|c|c c c|ccccccc|c|}
    \hline
        Group & Dataset & Resolution & \#images & \multicolumn{3}{c|}{Original Splitting} & \multicolumn{7}{c|}{Labels} & train/test \\
        \cline{5-14}
        & & & & Train & Valid & Test & N & D & G & C & A & H & P & \\
        \hline
        \multirow{6}{*}{Meta} & Kaggle-CAT\footnotemark[2]  & 2592 × 1728 or 2464×1632 & 600 & 360 & 120 & 120 & \checkmark &  & \checkmark & \checkmark &  &  &  & \checkmark/\checkmark \\
        & DDR \cite{Li2019} & max: 5184×3456 min: 512×512 & 12522 & 6261 & 2504 & 3757 &  & \checkmark &  &  &  &  &  & \checkmark/\checkmark \\
        & REFUGE2 \cite{Fang2022a} & max: 2124×2056 min: 1634×1634 & 2000 & 1200 & 400 & 400 &  &  & \checkmark &  &  &  &  & \checkmark/\checkmark \\
        & ADAM \cite{Fang2022} & 2124 × 2056 or 1444×1444 & 1200 & 400 & 400 & 400 &  &  &  &  & \checkmark &  &  & \checkmark/\checkmark \\
        & Kaggle-HR\footnotemark[3]  & 800×800 & 712 & 427 & 142 & 143 &  &  &  &  &  & \checkmark &  & \checkmark/\checkmark \\
        & PALM \cite{fang2024open} & 2124 × 2056 or 1444×1444 & 1200 & 400 & 400 & 400 &  &  &  &  &  &  & \checkmark & \checkmark/\checkmark \\
        \hline
        \multirow{4}{*}{Unseen} & RFMiD \cite{Pachade2021} & max: 4288×2848 min: 2048×1536 & 3200 & 1920 & 640 & 640 & \checkmark &  &  &  & \checkmark &  &  & \ding{55}/\checkmark \\
        & APTOS2019\footnotemark[4]  & max: 4288×2848 min: 474×358 & 3662 & — & — & — &  & \checkmark &  &  &  &  &  & \ding{55}/\checkmark \\
        & ORIGA$^{\text{light}}$ \cite{Zhang2010} & 3072 × 2048 & 650 & — & — & — &  &  & \checkmark &  &  &  &  & \ding{55}/\checkmark \\
        & HPMI \cite{Huang2023} & 512 × 512 & 4011 & — & — & — &  &  &  &  &  &  & \checkmark & \ding{55}/\checkmark \\
        \hline
        ODIR & ODIR \cite{OIA-ODIR-2021} & max: 5184×3456 min: 160x120 & 7949 & 6961 & — & 988 & \checkmark & \checkmark & \checkmark & \checkmark & \checkmark & \checkmark & \checkmark & \ding{55}/\checkmark \\
        ODIR 200×3 & ODIR 200x3 \cite{SilvaRodriguez2025} & max: 5184×3456 min: 868x793 & 600 & — & — & — & \checkmark &  &  & \checkmark &  &  & \checkmark & \ding{55}/\checkmark \\
        \hline
    \end{tabular}}
    \caption{Details of open-access datasets used in this study. Here, 'N' is the normal state, 'D' is diabetic retinopathy (DR), 'G' is glaucoma, 'C' is cataract, 'A' is age-related macular degeneration (AMD), 'H' is hypertensive retinopathy (HR), and 'P' is pathologic myopia (PM).}
    \label{table1:dataset_info}
\end{table*}

\footnotetext[1]{\url{https://odir2019.grand-challenge.org/Download/}}
\footnotetext[2]{\label{fn:CAT} \url{https://www.kaggle.com/datasets/jr2ngb/cataractdataset}}
\footnotetext[3]{\url{https://www.kaggle.com/datasets/harshwardhanfartale/hypertension-and-hypertensive-retinopathy-dataset}}
\footnotetext[4]{\url{https://www.kaggle.com/competitions/aptos2019-blindness-detection/data}}

\begin{table*}[!t]
\centering
\Large
\resizebox{\textwidth}{!}{
\begin{tabular}{|l|cccccccccc|}
\hline
Methods             & T1: Normal & T2: DR & \multicolumn{3}{c}{T3: Glaucoma}                            & T4: Cataract & T5: AMD & T6: HR & T7: PM & $mF$ \\ 
\cline{4-6}
                              & Kaggle-CAT       & DDR                      & REFUGE2                  & Kaggle-CAT          & Average & Kaggle-CAT          & ADAM               & Kaggle-HR & PALM              &                                    \\ 
\hline
MultiNets                     & 78.1$_{\pm1.2}$  & \textbf{67.8$_{\pm0.9}$}   & 75.7$_{\pm1.3}$   & \textbf{89.5$_{\pm1.0}$}   & \underline{82.6$_{\pm1.0}$}   & 88.0$_{\pm2.0}$   & 85.3$_{\pm1.7}$   & 80.4$_{\pm3.7}$   & \textbf{96.7$_{\pm0.4}$}   & 82.7$_{\pm0.7}$          \\ 

MultiHeads                    & 86.8$_{\pm0.8}$   & 66.8$_{\pm1.0}$   & \underline{77.2$_{\pm1.4}$}   & 85.4$_{\pm0.4}$   & 81.3$_{\pm0.6}$   & 88.8$_{\pm3.2}$   & 84.4$_{\pm1.2}$   & 81.1$_{\pm1.7}$   & 96.3$_{\pm0.1}$   & 83.6$_{\pm0.7}$          \\  

SST$_{AAAI2022}$~\cite{chen2022}                            & \textbf{87.6$_{\pm0.8}$}   & \underline{67.1$_{\pm1.1}$}   & 73.6$_{\pm1.1}$   & 87.3$_{\pm1.5}$   & 80.5$_{\pm0.6}$   & 92.3$_{\pm2.8}$   & 83.7$_{\pm2.2}$   & 77.6$_{\pm3.4}$   & 95.2$_{\pm0.7}$   & 83.4$_{\pm1.4}$          \\ 

SARB$_{AAAI2022}$~\cite{pu2022semantic}                           & 86.2$_{\pm3.3}$ & 66.7$_{\pm1.6}$ & 77.0$_{\pm2.4}$ & 86.9$_{\pm0.7}$ & 81.9$_{\pm1.5}$ & 90.0$_{\pm0.9}$ & \textbf{85.9$_{\pm0.9}$} & 81.4$_{\pm2.1}$ & 96.0$_{\pm1.0}$ & \underline{84.0$_{\pm0.9}$} \\

BoostLU$_{CVPR2023}$~\cite{kim2023}                             & 80.2$_{\pm4.2}$   & 62.0$_{\pm1.3}$   & 72.6$_{\pm1.8}$   & 85.8$_{\pm3.1}$   & 78.6$_{\pm2.6}$   & 90.0$_{\pm2.0}$   & 76.8$_{\pm2.3}$   & 78.8$_{\pm1.5}$   & 95.9$_{\pm0.4}$   & 80.3$_{\pm0.5}$          \\ 

HST$_{IJCV2024}$~\cite{Chen2024}                          & \underline{87.0$_{\pm0.5}$}   & 67.0$_{\pm0.3}$   & 72.3$_{\pm2.5}$   & \underline{89.0$_{\pm3.1}$}   & 80.7$_{\pm0.8}$   & \underline{92.6$_{\pm1.7}$}   & 81.2$_{\pm2.1}$   & 79.6$_{\pm1.0}$   & 95.6$_{\pm1.0}$   & 83.4$_{\pm0.4}$          \\ 

CALDNR$_{TMM2024}$~\cite{Pu2024}                         & 82.5$_{\pm3.1}$   & 63.8$_{\pm1.3}$   & 74.6$_{\pm4.0}$   & 88.6$_{\pm1.6}$   & 81.6$_{\pm1.4}$   & \textbf{93.1$_{\pm1.5}$}   & 75.6$_{\pm7.3}$   & \underline{81.8$_{\pm3.8}$}   & 95.7$_{\pm0.2}$   & 82.0$_{\pm1.1}$          \\ 

PSScreen$_{Ours}$  & 86.4$_{\pm1.7}$  & 64.7$_{\pm1.7}$  & \textbf{77.7$_{\pm1.5}$} & \underline{89.0$_{\pm3.1}$}  & \textbf{83.3$_{\pm2.0}$}  & 89.1$_{\pm2.4}$  & \underline{84.5$_{\pm1.4}$}  & \textbf{85.0$_{\pm1.5}$}  & \underline{96.5$_{\pm0.4}$}  & \textbf{84.2$_{\pm0.3}$}  \\ 
\hline
\end{tabular}}
\caption{Comparison of F-score for each disease on the meta-dataset. The best and second-best are highlighted in bold and with an underline. Means and standard deviations are reported over three trials.}
\label{tab:f detail comparison in Meta}
\end{table*}

\begin{table*}[!t]
\centering
\Large
\resizebox{\textwidth}{!}{
\begin{tabular}{|l|cccccccccc|}
\hline
Methods             & T1: Normal & T2: DR & \multicolumn{3}{c}{T3: Glaucoma}                            & T4: Cataract & T5: AMD & T6: HR & T7: PM & $mQWK$ \\ 
\cline{4-6}
                              & Kaggle-CAT       & DDR                      & REFUGE2                  & Kaggle-CAT         & Average & Kaggle-CAT         & ADAM               & Kaggle-HR & PALM              &                                    \\ 
\hline
MultiNets                     & 56.5$_{\pm2.6}$   & 88.1$_{\pm0.1}$   & 51.4$_{\pm2.6}$   & \textbf{79.0$_{\pm4.0}$}   & \underline{65.2$_{\pm3.7}$}   & 76.0$_{\pm4.1}$   & 70.5$_{\pm3.5}$   & 61.2$_{\pm7.4}$   & \textbf{93.5$_{\pm0.9}$}   & 73.0$_{\pm1.4}$          \\ 

MultiHeads                    & 73.6$_{\pm1.6}$   & 87.8$_{\pm0.8}$   & \underline{54.5$_{\pm2.8}$}   & 71.0$_{\pm0.7}$   & 62.7$_{\pm1.2}$   & 77.5$_{\pm6.5}$   & 68.7$_{\pm2.5}$   & 62.3$_{\pm3.5}$   & 92.6$_{\pm0.3}$   & 75.0$_{\pm1.3}$          \\ 

SST$_{AAAI2022}$~\cite{chen2022}                            & \textbf{75.2$_{\pm1.6}$}   & \underline{88.4$_{\pm0.2}$}   & 47.6$_{\pm2.1}$   & 74.6$_{\pm3.1}$   & 61.1$_{\pm1.4}$   & 84.7$_{\pm5.5}$   & 67.5$_{\pm4.2}$   & 55.2$_{\pm6.8}$   & 90.4$_{\pm1.3}$   & 74.7$_{\pm2.8}$          \\ 

SARB$_{AAAI2022}$~\cite{pu2022semantic}                          & 72.5$_{\pm6.7}$   & 87.3$_{\pm0.9}$   & 54.0$_{\pm4.7}$   & 77.1$_{\pm1.5}$   & 63.9$_{\pm3.0}$   & 80.0$_{\pm1.8}$   & \textbf{71.7$_{\pm1.7}$}   & 62.7$_{\pm4.3}$   & 92.0$_{\pm0.9}$   & \underline{75.7$_{\pm1.8}$}          \\ 

BoostLU$_{CVPR2023}$~\cite{kim2023}                           & 60.8$_{\pm8.2}$   & 84.0$_{\pm2.2}$   & 46.1$_{\pm3.6}$   & 71.7$_{\pm6.0}$   & 57.6$_{\pm5.2}$   & 80.1$_{\pm4.0}$   & 54.7$_{\pm4.1}$   & 57.6$_{\pm3.0}$   & 91.8$_{\pm0.8}$   & 69.5$_{\pm1.1}$          \\ 

HST$_{IJCV2024}$~\cite{Chen2024}                           & \underline{74.1$_{\pm0.9}$}   & \textbf{88.8$_{\pm0.1}$}   & 45.0$_{\pm4.5}$   & 78.0$_{\pm6.2}$   & 61.5$_{\pm1.6}$   & \underline{85.3$_{\pm3.5}$}   & 62.5$_{\pm4.3}$   & 59.3$_{\pm1.8}$   & 91.3$_{\pm1.0}$   & 74.7$_{\pm0.8}$          \\ 

CALDNR$_{TMM2024}$~\cite{Pu2024}                        & 65.3$_{\pm5.9}$   & 87.0$_{\pm0.8}$   & 50.0$_{\pm7.3}$   & 77.2$_{\pm3.2}$   & 63.6$_{\pm2.4}$   & \textbf{86.3$_{\pm3.1}$}   & 52.4$_{\pm13.2}$  & \underline{63.7$_{\pm7.5}$}   & 91.5$_{\pm0.5}$   & 72.8$_{\pm2.2}$          \\ 

PSScreen$_{Ours}$   & 73.0$_{\pm3.4}$  & 87.4$_{\pm0.2}$  & \textbf{55.4$_{\pm3.0}$}  & \underline{78.1$_{\pm6.2}$}  & \textbf{66.8$_{\pm4.0}$}  & 78.3$_{\pm4.7}$  & \underline{69.1$_{\pm2.7}$}  & \textbf{70.1$_{\pm3.0}$}  & \underline{93.0$_{\pm0.9}$}  & \textbf{76.8$_{\pm0.8}$}  \\ 
\hline
\end{tabular}}
\caption{Comparison of QWK for each disease on the meta-dataset. The best and second-best are highlighted in bold and with an underline. Means and standard deviations are reported over three trials.}
\label{tab:qwk detail comparison in Meta}
\end{table*}

\section{Experimental Results for Each Disease}
\label{sppl:class-wise-rst}

\subsection{Performances for Each Disease on the Meta-dataset}

We separately report $mF$ and $mQWK$ of PSScreen, the baselines, and the existing SOTAs for each disease on the meta-dataset in Table~\ref{tab:f detail comparison in Meta} and Table~\ref{tab:qwk detail comparison in Meta} respectively. It can be observed that our PSScreen outperforms the second-best SARB \cite{pu2022semantic} by 0.2\% in $mF$ and 1.1\% in $mQWK$. More detailed, PSScreen achieves the best or second-best in detecting glaucoma, AMD, HR, and PM. Despite the significant domain shift between the training and test sets of REFUGE2, our PSScreen outperforms the second-best MultiHeads on REFUGE2 by 0.9\% in QWK, further demonstrating the good generalization capability of our model.


\begin{table*}[!h]
    \centering
    \large
    \resizebox{\textwidth}{!}{
    \begin{tabular}{|l|cc|cc|cc|cc|cc|cc|}
        \hline
         Methods & \multicolumn{2}{c|}{T1: Normal } & \multicolumn{2}{c|}{T2: DR} & \multicolumn{2}{c|}{T3: Glaucoma} & \multicolumn{2}{c|}{T4: AMD} & \multicolumn{2}{c|}{T5: PM} & \multicolumn{2}{c|}{Average} \\
         \cline{2-11}
         & \multicolumn{2}{c|}{RFMID} & \multicolumn{2}{c|}{APTOS} & \multicolumn{2}{c|}{ORIGA} & \multicolumn{2}{c|}{RFMID} & \multicolumn{2}{c|}{HPMI} & \multicolumn{2}{c|}{} \\
         \cline{2-11}
        & F-score & QWK & F-score & QWK & F-score & QWK & F-score & QWK & F-score & QWK & $mF$ & $mQWK$ \\
        \hline
        MultiNets    & 68.1$_{\pm2.3}$ & 36.7$_{\pm4.2}$ & 41.3$_{\pm2.2}$ & 70.2$_{\pm5.8}$ & \textbf{71.3$_{\pm0.7}$} & \textbf{42.8$_{\pm1.5}$} & 48.3$_{\pm2.4}$ & 7.3$_{\pm1.1}$ & \textbf{83.1$_{\pm1.1}$} & \textbf{66.8$_{\pm2.1}$} & 62.4$_{\pm1.2}$ & 44.8$_{\pm2.3}$ \\
        MultiHeads   & 67.4$_{\pm4.1}$ & 36.3$_{\pm7.5}$ & \underline{44.7$_{\pm2.7}$} & \underline{77.7$_{\pm4.7}$} & 67.9$_{\pm1.8}$ & 36.7$_{\pm3.0}$ & 52.3$_{\pm2.0}$ & 9.9$_{\pm2.1}$ & 79.6$_{\pm0.4}$ & 60.4$_{\pm0.8}$ & 62.4$_{\pm1.5}$ & 44.2$_{\pm2.0}$ \\
        SST$_{AAAI2022}$~\cite{chen2022}   & \underline{76.8$_{\pm4.2}$} & \underline{53.6$_{\pm8.5}$} & 42.0$_{\pm0.5}$ & 74.9$_{\pm2.1}$ & 63.9$_{\pm1.6}$ & 29.0$_{\pm2.7}$ & 52.1$_{\pm1.3}$ & 12.4$_{\pm2.6}$ & \underline{81.3$_{\pm1.0}$} & \underline{63.5$_{\pm1.1}$} & \underline{63.2$_{\pm0.9}$} & \underline{46.7$_{\pm1.8}$} \\
        SARB$_{AAAI2022}$~\cite{pu2022semantic}  & 63.4$_{\pm4.3}$ & 28.2$_{\pm7.4}$ & 41.7$_{\pm0.8}$ & 70.8$_{\pm3.3}$ & 65.8$_{\pm0.7}$ & 32.0$_{\pm1.8}$ & \underline{57.4$_{\pm3.3}$} & \underline{18.9$_{\pm2.8}$} & 74.3$_{\pm2.6}$ & 51.3$_{\pm4.2}$ & 60.5$_{\pm1.0}$ & 40.3$_{\pm1.2}$ \\
        BoostLU$_{CVPR2023}$~\cite{kim2023}   & 46.0$_{\pm1.2}$ & 1.2$_{\pm1.5}$ & 37.3$_{\pm1.2}$ & 58.1$_{\pm2.9}$ & 52.8$_{\pm1.9}$ & 13.7$_{\pm2.7}$ & 53.9$_{\pm1.8}$ & 10.3$_{\pm3.5}$ & 32.6$_{\pm3.8}$ & 7.2$_{\pm1.2}$ & 44.5$_{\pm0.8}$ & 29.8$_{\pm2.3}$ \\
        HST$_{IJCV2024}$~\cite{Chen2024}   & 74.6$_{\pm5.9}$ & 48.4$_{\pm11.5}$ & 42.9$_{\pm2.3}$ & 74.0$_{\pm5.9}$ & 64.7$_{\pm2.0}$ & 29.7$_{\pm3.8}$ & 50.9$_{\pm0.7}$ & 18.0$_{\pm0.6}$ & 78.8$_{\pm1.8}$ & 59.0$_{\pm3.2}$ & 62.3$_{\pm1.7}$ & 44.8$_{\pm3.8}$ \\
        CALDNR$_{TMM2024}$~\cite{Pu2024}   & 55.8$_{\pm4.4}$ & 14.9$_{\pm9.7}$ & 32.1$_{\pm6.3}$ & 42.1$_{\pm18.5}$ & 58.8$_{\pm1.1}$ & 22.3$_{\pm1.7}$ & 50.6$_{\pm3.2}$ & 3.3$_{\pm5.7}$ & 41.0$_{\pm3.7}$ & 12.5$_{\pm6.2}$ & 47.6$_{\pm3.1}$ & 19.0$_{\pm6.7}$ \\
        PSScreen$_{Ours}$   & \textbf{80.2$_{\pm1.4}$} & \textbf{60.5$_{\pm2.5}$} & \textbf{46.7$_{\pm0.6}$} & \textbf{84.5$_{\pm0.9}$} & \underline{68.0$_{\pm1.2}$} & \underline{36.9$_{\pm2.2}$} & \textbf{59.1$_{\pm2.6}$} & \textbf{19.6$_{\pm4.6}$} & 75.4$_{\pm0.8}$ & 53.1$_{\pm1.3}$ & \textbf{65.9$_{\pm0.1}$} & \textbf{50.9$_{\pm0.1}$} \\
        \hline
    \end{tabular}}
    \caption{Comparison of F-score and QWK for each disease on the unseen-dataset. The best and second-best are highlighted in bold and with an underline. Means and standard deviations are reported over three trials.}
    \label{tab:metrics comparison in Unseen}
\end{table*}

\subsection{Performances for Each Disease on the Unseen and ODIR Datasets}

We present the performances for each disease on unseen-dataset in Table~\ref{tab:metrics comparison in Unseen}. As shown, PSScreen outperforms the second-best SST \cite{chen2022} by 2.7\% in $mF$ and 4.2\% in $mQWK$. More specifically, PSScreen achieves the best performance in the normal state detection, AMD detection, and DR grading tasks. Notably, it yields substantial improvements in QWK, outperforming the second-best SST \cite{chen2022} by 6.9\% in the normal state detection and the second-best MultiHeads by 6.8\% in DR grading, respectively. Additionally, in the glaucoma detection task, PSScreen ranks second. 

We also report the F-score and QWK on ODIR dataset \cite{OIA-ODIR-2021} in Table~\ref{tab:f detail comparison in ODIR} and~\ref{tab:qwk detail comparison in ODIR} respectively. As a reference for the upper bound performances, the performances by the fully supervised model are also reported in the first row of table~\ref{tab:f detail comparison in ODIR} and table~\ref{tab:qwk detail comparison in ODIR}. It can be observed that PSScreen achieves a significant improvement over the second-best SARB \cite{pu2022semantic}, with $mF$ improving by 2.7\% and $mQWK$ improving by 6.0\%. More detailed, we find that PSScreen achieves the best performance in all tasks except for HR detection. Notably, for the normal state detection, DR grading, and glaucoma detection, PSScreen demonstrates a significant improvement in the QWK metric over the respective second-best approaches SST \cite{chen2022}, BoostLU \cite{kim2023}, and CALDNR \cite{Pu2024}—with gains of 5.7\%, 5.2\%, and 6.7\%, respectively. Especially for glaucoma detection and DR grading, our model's F-score even surpasses that of the fully supervised model, further validating the strong domain generalization capability of PSScreen. Additionally, for detecting HR, all methods perform poorly, and even the fully supervised model struggles to make accurate predictions. 

\begin{table*}[!t]
\centering
\Large
\resizebox{0.7\textwidth}{!}{
\begin{tabular}{|l|ccccccc|c|}
\hline
Methods             & T1: Normal & T2: DR & T3: Glaucoma & T4: Cataract & T5: AMD & T6: HR & T7: PM & $mF$ \\ 
\hline
Full supervise          & 70.8$_{\pm 0.2}$  & 32.1$_{\pm 1.7}$       & 66.8$_{\pm 0.9}$     & 91.8$_{\pm 0.8}$     & 80.2$_{\pm 1.4}$  & 58.8$_{\pm 1.1}$  & 87.8$_{\pm 1.2}$  & 69.8$_{\pm 0.2}$           \\  

MultiNets           & 59.3$_{\pm 2.6}$  & 28.7$_{\pm 0.7}$       & 60.8$_{\pm 1.7}$     & 66.2$_{\pm 1.9}$     & 59.2$_{\pm 5.7}$  & 47.7$_{\pm 2.9}$  & 72.7$_{\pm 1.4}$  & 56.4$_{\pm 1.4}$            \\ 

MultiHeads          & 53.9$_{\pm 1.1}$  & 29.7$_{\pm 1.3}$       & 62.8$_{\pm 2.4}$     & 67.3$_{\pm 0.5}$     & 57.4$_{\pm 2.4}$  & 47.7$_{\pm 1.1}$  & 78.2$_{\pm 1.9}$  & 56.7$_{\pm 0.6}$            \\ 

SST$_{AAAI2022}$~\cite{chen2022}                  & \underline{62.8$_{\pm 3.2}$}  & 29.1$_{\pm 1.6}$       & 58.7$_{\pm 1.0}$     & 70.9$_{\pm 3.9}$     & 53.0$_{\pm 1.1}$  & 44.1$_{\pm 1.5}$  & 70.6$_{\pm 4.2}$  & 55.6$_{\pm 1.7}$            \\ 

SARB$_{AAAI2022}$~\cite{pu2022semantic}                 & 60.8$_{\pm 2.2}$  & 31.7$_{\pm 2.3}$       & 59.1$_{\pm 1.9}$     & \underline{84.6$_{\pm 2.2}$}     & 61.2$_{\pm 4.2}$  & \underline{51.0$_{\pm 1.3}$}  & \underline{81.3$_{\pm 0.9}$}  & \underline{61.4$_{\pm 0.9}$}            \\ 
BoostLU$_{CVPR2023}$~\cite{kim2023}                  & 36.1$_{\pm 0.0}$  & 29.6$_{\pm 2.8}$       & \underline{63.5$_{\pm 3.4}$}     & 48.8$_{\pm 0.0}$     & \underline{61.7$_{\pm 5.5}$}  & 49.0$_{\pm 0.0}$  & 66.8$_{\pm 3.1}$  & 50.6$_{\pm 1.0}$            \\ 
HST$_{IJCV2024}$~\cite{Chen2024}                 & 60.9$_{\pm 2.4}$  & 29.0$_{\pm 1.9}$       & 60.4$_{\pm 1.2}$     & 76.0$_{\pm 2.8}$     & 50.4$_{\pm 2.0}$  & 43.2$_{\pm 1.7}$  & 71.9$_{\pm 1.7}$  & 56.0$_{\pm 0.8}$            \\ 

CALDNR$_{TMM2024}$~\cite{Pu2024}               & 36.1$_{\pm 0.0}$  & \textbf{33.5$_{\pm 3.1}$}       & \underline{63.5$_{\pm 3.4}$}     & 55.9$_{\pm 10.6}$    & 58.9$_{\pm 3.7}$  & 49.3$_{\pm 0.0}$  & 67.9$_{\pm 1.8}$  & 52.1$_{\pm 0.6}$            \\

PSScreen$_{Ours}$  & \textbf{66.8$_{\pm 1.4}$}  & \underline{33.1$_{\pm 1.9}$}       & \textbf{66.9$_{\pm 2.3}$}     & \textbf{85.9$_{\pm 1.6}$}     & \textbf{63.0$_{\pm 0.8}$}  & \textbf{51.1$_{\pm 2.3}$}  & \textbf{81.6$_{\pm 2.6}$}  & \textbf{64.1$_{\pm 1.0}$}            \\ 

\hline
\end{tabular}}
\caption{Comparison of F-score for each disease on the ODIR dataset. The best and second-best are highlighted in bold and with an underline. Means and standard deviations are reported over three trials.}
\label{tab:f detail comparison in ODIR}
\end{table*}

\begin{table*}[!t]
\centering
\Large
\resizebox{0.7\textwidth}{!}{
\begin{tabular}{|l|ccccccc|c|}
\hline
Methods            & T1: Normal & T2: DR & T3: Glaucoma & T4: Cataract & T5: AMD & T6: HR & T7: PM & $mQWK$ \\
\hline

Full supervise         & 41.6$_{\pm 0.4}$  & 53.6$_{\pm 3.1}$       & 33.8$_{\pm 1.7}$     & 83.5$_{\pm 1.5}$     & 60.5$_{\pm 2.8}$  & 19.2$_{\pm 0.8}$  & 75.6$_{\pm 2.3}$  & 52.6$_{\pm 0.7}$            \\ 

MultiNets           & 19.2$_{\pm 4.4}$  & 29.7$_{\pm 1.6}$       & 24.8$_{\pm 2.2}$     & 35.4$_{\pm 3.2}$     & 22.9$_{\pm 8.3}$  & \textbf{3.9$_{\pm 2.5}$}   & 46.6$_{\pm 2.5}$  & 26.1$_{\pm 2.1}$            \\ 

MultiHeads          & 17.9$_{\pm 1.9}$  & 34.0$_{\pm 2.5}$       & 26.9$_{\pm 4.1}$     & 36.5$_{\pm 1.3}$     & 18.8$_{\pm 3.6}$  & \underline{3.2$_{\pm 1.7}$}   & 56.8$_{\pm 3.6}$  & 27.7$_{\pm 0.4}$            \\ 

SST$_{AAAI2022}$~\cite{chen2022}                   & \underline{28.9$_{\pm 4.3}$}  & 32.3$_{\pm 1.0}$       & 21.7$_{\pm 1.5}$     & 43.3$_{\pm 6.8}$     & 14.0$_{\pm 1.8}$  & 0.3$_{\pm 1.1}$   & 42.8$_{\pm 7.5}$  & 26.2$_{\pm 2.5}$            \\

SARB$_{AAAI2022}$~\cite{pu2022semantic}                 & 24.3$_{\pm 3.6}$  & 31.9$_{\pm 2.1}$       & 21.0$_{\pm 2.9}$     & \underline{69.3$_{\pm 4.5}$}     & \underline{24.5$_{\pm 7.2}$}  & 2.6$_{\pm 2.2}$  & \underline{62.9$_{\pm 1.8}$}  & \underline{33.8$_{\pm 1.6}$}            \\ 

BoostLU$_{CVPR2023}$~\cite{kim2023}                    & 0.0$_{\pm 0.0}$   & \underline{39.6$_{\pm 7.3}$}       & 24.7$_{\pm 1.0}$     & 0.0$_{\pm 0.0}$      & \underline{24.5$_{\pm 10.7}$} & 0.0$_{\pm 0.0}$   & 34.0$_{\pm 6.2}$  & 17.5$_{\pm 2.1}$            \\ 

HST$_{IJCV2024}$~\cite{Chen2024}                    & 25.4$_{\pm 2.9}$  & 33.1$_{\pm 0.4}$       & 23.9$_{\pm 1.5}$     & 52.7$_{\pm 5.2}$     & 11.8$_{\pm 1.9}$  & 0.6$_{\pm 0.4}$   & 45.3$_{\pm 3.2}$  & 27.5$_{\pm 0.8}$            \\ 

CALDNR$_{TMM2024}$~\cite{Pu2024}                 & 0.0$_{\pm 0.0}$   & 38.7$_{\pm 3.9}$       & \underline{27.3$_{\pm 6.6}$}     & 13.5$_{\pm 20.3}$    & 18.7$_{\pm 7.1}$  & 0.0$_{\pm 0.0}$   & 36.1$_{\pm 3.6}$  & 19.1$_{\pm 0.8}$            \\ 

PSScreen$_{Ours}$  & \textbf{34.6$_{\pm 2.1}$}  & \textbf{44.8$_{\pm 2.5}$}       & \textbf{34.0$_{\pm 4.4}$}     & \textbf{71.9$_{\pm 3.1}$}     & \textbf{27.6$_{\pm 1.4}$}  & 2.3$_{\pm 4.6}$   & \textbf{63.4$_{\pm 5.2}$}  & \textbf{39.8$_{\pm 1.3}$}            \\ 

\hline
\end{tabular}}
\caption{Comparison of QWK for each disease on the ODIR dataset. The best and second-best are highlighted in bold and with an underline. Means and standard deviations are reported over three trials.}
\label{tab:qwk detail comparison in ODIR}
\end{table*}

\section{Additional Ablation Study Results}

\subsection{Influences of Loss Weights $\lambda_1$, $\lambda_2$, and $\lambda_3$}
\label{sppl:ablation_loss}
We vary $\lambda_1$, $\lambda_2$, and $\lambda_3$, and report $mF$ in Table \ref{tab:backbone_mu_supp}(a), which shows that $\lambda_1=0.05$, $\lambda_2=1.0$, and $\lambda_3=0.6$ perform the best performances.

\subsection{Compatibility with Various Backbones}
\label{sppl:ablation_backbone}
To further validate the compatibility of our PSScreen, we report $mF$ of PSScreen with various backbones including ConvNeXt-T \cite{Liu2022}, ConvNeXt V2-T \cite{Woo2023}, Swin-T \cite{Liu2021}, and VMamba-T \cite{Liu2024d} and compare them with the naive ones, i.e., MultiHeads with various backbones in Table~\ref{tab:backbone_mu_supp}(b). The results demonstrate that our PSScreen consistently improves the performances across all three datasets with various backbones. 

\begin{table}[t!]
\centering
\begin{minipage}[t]{0.49\textwidth}
    \renewcommand{\arraystretch}{1.03} 
    \centering
    \resizebox{\textwidth}{!}{
    \begin{tabular}{|c|c|c|c|c|c|c|}
        \hline
        $\lambda_1$ & $\lambda_2$ & $\lambda_3$ & Meta & Unseen & ODIR \\
        \hline
        0.1   & 1.0   & 0.6 & 83.2\scriptsize$\pm$0.2 & 65.4\scriptsize$\pm$0.1 & 63.6\scriptsize$\pm$0.5 \\
        \textbf{0.05}  & \textbf{1.0}   & \textbf{0.6} & \textbf{84.2\scriptsize$\pm$0.3} & \textbf{65.9\scriptsize$\pm$0.1} & \textbf{64.1\scriptsize$\pm$1.0} \\
        0.025 & 1.0   & 0.6 & 84.1\scriptsize$\pm$0.1 & 65.7\scriptsize$\pm$0.4 & 63.9\scriptsize$\pm$0.5 \\
        \hline
        0.05  & 0.5   & 0.6 & 83.2\scriptsize$\pm$0.1 & 65.2\scriptsize$\pm$0.2 & 63.4\scriptsize$\pm$0.4 \\
        \textbf{0.05}  & \textbf{1.0}   & \textbf{0.6} & \textbf{84.2\scriptsize$\pm$0.3} & \textbf{65.9\scriptsize$\pm$0.1} & \textbf{64.1\scriptsize$\pm$1.0} \\
        0.05  & 2.0   & 0.6 & 82.9\scriptsize$\pm$0.2 & 64.5\scriptsize$\pm$0.8 & 63.6\scriptsize$\pm$0.2 \\
        \hline
        0.05  & 1.0   & 0.4 & 84.1\scriptsize$\pm$0.5 & 64.7\scriptsize$\pm$0.6 & 63.9\scriptsize$\pm$0.4 \\
        \textbf{0.05}  & \textbf{1.0}   & \textbf{0.6} & \textbf{84.2\scriptsize$\pm$0.3} & \textbf{65.9\scriptsize$\pm$0.1} & \textbf{64.1\scriptsize$\pm$1.0} \\
        0.05  & 1.0   & 0.8 & 84.2\scriptsize$\pm$0.3 & 65.4\scriptsize$\pm$0.9 & 63.2\scriptsize$\pm$0.1 \\
        \hline
        \end{tabular}
    }
    {\vspace{2pt} \small (a) F-score under different loss weight settings.}
\end{minipage}
\begin{minipage}[t]{0.48\textwidth}
    \centering
    \resizebox{\textwidth}{!}{
     \begin{tabular}{|l|rrr|}
        \hline
        Methods & Meta & Unseen & ODIR
        \\ \hline
        ResNet-101 \cite{He2016}        &     83.6\scriptsize$\pm$0.7               &      62.4\scriptsize$\pm$1.5               &  56.7\scriptsize$\pm$0.6     \\             
        +PSScreen              &       \textbf{84.2\scriptsize$\pm$0.3}              &      \textbf{65.9\scriptsize$\pm$0.1}                   &       \textbf{64.1\scriptsize$\pm$1.0}                \\ \hline
        ConvNeXt-T \cite{Liu2022}      &        83.6\scriptsize$\pm$0.3             &        64.4\scriptsize$\pm$0.6              &    55.8\scriptsize$\pm$1.8   \\               
        +PSScreen                &         \textbf{84.5\scriptsize$\pm$0.2}            &      \textbf{65.6\scriptsize$\pm$0.7}                     &              \textbf{62.9\scriptsize$\pm$0.8}       \\ \hline
        ConvNeXt V2-T \cite{Woo2023}          &      84.4\scriptsize$\pm$0.4               &         65.1\scriptsize$\pm$0.6                   &   58.1\scriptsize$\pm$0.9     \\              
        +PSScreen                   &     \textbf{84.5\scriptsize$\pm$0.1}                &       \textbf{65.3\scriptsize$\pm$0.5}               &        \textbf{63.7\scriptsize$\pm$0.7}               \\ \hline
        Swin-T \cite{Liu2021}          &      83.6\scriptsize$\pm$0.2               &         65.5\scriptsize$\pm$0.5                   &   56.1\scriptsize$\pm$0.3     \\              
        +PSScreen                   &     \textbf{85.3\scriptsize$\pm$0.5}                &       \textbf{66.2\scriptsize$\pm$0.3}               &        \textbf{61.8\scriptsize$\pm$0.5}               \\ \hline
        
        VMamba-T  \cite{Liu2024d}           &      84.0\scriptsize$\pm$0.8               &         64.3\scriptsize$\pm$0.4                   &   56.8\scriptsize$\pm$0.4     \\              
        +PSScreen                   &     \textbf{84.9\scriptsize$\pm$0.1}                &       \textbf{65.3\scriptsize$\pm$1.3}               &        \textbf{62.6\scriptsize$\pm$1.0}               \\
        \hline
    \end{tabular}
    }
    {\vspace{2.5pt} \small (b) Performances with different backbones.}
\end{minipage}

\caption{(a) $mF$ under different settings of $\lambda_1$, $\lambda_2$, and $\lambda_3$. (b) $mF$ by MultiHeads and PSScreen with different backbones.}
\label{tab:backbone_mu_supp}
\end{table}

\begin{table}[t!]
   \centering
   \resizebox{.6\columnwidth}{!}{ 
   \begin{tabular}{|l|cc|cc|cc|}
       \hline
       \multirow{2}{*}{$\tau$} & \multicolumn{2}{c|}{Meta} & \multicolumn{2}{c|}{Unseen} & \multicolumn{2}{c|}{ODIR} \\
       \cline{2-7}
                                & $mF$ & $mQWK$ & $mF$ & $mQWK$ & $mF$ & $mQWK$ \\
       \hline
       0.99   & 83.9\scriptsize$\pm$0.2 &76.1\scriptsize$\pm$0.4 & 64.7\scriptsize$\pm$0.3 & 49.4\scriptsize$\pm$0.5 & 62.9\scriptsize$\pm$0.3 & 38.7\scriptsize$\pm$0.6 \\
       0.95 & \textbf{84.2\scriptsize$\pm$0.3} & \textbf{76.8\scriptsize$\pm$0.8} & \textbf{65.9\scriptsize$\pm$0.1}  & \textbf{50.9\scriptsize$\pm$0.1} & \textbf{64.1\scriptsize$\pm$1.0} & \textbf{39.8\scriptsize$\pm$1.3} \\
        0.90    & 84.0\scriptsize$\pm$0.3 &76.6\scriptsize$\pm$0.5 & 65.0\scriptsize$\pm$0.3 & 50.1\scriptsize$\pm$0.5 & 63.7\scriptsize$\pm$0.2 & 39.3\scriptsize$\pm$0.3 \\
       0.85  & 83.8\scriptsize$\pm$0.1 & 76.1\scriptsize$\pm$0.1 & 64.4\scriptsize$\pm$0.2  & 48.8\scriptsize$\pm$0.4 & 63.0\scriptsize$\pm$0.8 & 38.0\scriptsize$\pm$1.2 \\
       \hline
   \end{tabular}
   }
    \caption{Performance analysis under different settings of $\tau$ on $mF$ and $mQWK$.}
    \label{tab: threshold tau}
\end{table}

\begin{table}[h!]
\centering
\renewcommand{\arraystretch}{0.8}
\begin{tabularx}{\textwidth}{|l|X|}
\hline
\textbf{Disease} & \textbf{Description} \\
\hline
normal & "healthy", "no findings", "no lesion signs", "no glaucoma", "no retinopathy" \\
\hline
glaucoma & "optic nerve abnormalities", "abnormal size of the optic cup", "anomalous size in the optic disc" \\
\hline
cataract & "opacity in the macular area" \\
\hline
age-related macular degeneration & "many small drusen", "few medium-sized drusen", "large drusen", "macular degeneration" \\
\hline
hypertensive retinopathy & "possible signs of haemorraghe with blot, dot, or flame-shaped", "possible presence of microaneurysm, cotton-wool spot, or hard exudate", "arteriolar narrowing", "vascular wall changes", "optic disk edema" \\
\hline
pathologic myopia & "anomalous disc, macular atrophy and possible tessellation" \\
\hline
no diabetic retinopathy & "no diabetic retinopathy", "no microaneurysms" \\
\hline
mild diabetic retinopathy & "only few microaneurysms" \\
\hline
moderate diabetic retinopathy & "many exudates near the macula", "many haemorrhages near the macula", "retinal thickening near the macula", "hard exudates", "cotton wool spots", "few severe haemorrhages" \\
\hline
severe diabetic retinopathy & "venous beading", "many severe haemorrhages", "intraretinal microvascular abnormality" \\
\hline
proliferative diabetic retinopathy & "preretinal or vitreous haemorrhage", "neovascularization" \\
\hline
\end{tabularx}
\caption{Expert knowledge descriptions for each retinal disease, which are directly borrowed from \cite{SilvaRodriguez2025}.}
\label{tab:expert describe}
\end{table}

\subsection{Influences of Threshold $\tau$ for Pseudo Label Generation}
$\tau$ in Eq. \ref{sd_unlabel_loss} is the threshold to obtain the pseudo labels and we vary its values and report the performances in Table~\ref{tab: threshold tau} to illustrate its effect. We observe that when $\tau$ is set too large, few samples satisfy the thresholding condition, limiting the supervision signal; whereas when 
$\tau$ is too small, many samples are incorrectly assigned pseudo labels, which harms the training stability and degrade performance. PSScreen achieves the best performance when $\tau$=0.95; therefore, we set $\tau$=0.95 in our experiments. Notably, PSScreen remains robust within a certain range of $\tau$ values, and only when the threshold is set excessively low, thereby resulting in a large number of noisy pseudo labels, does the performance exhibit a marked decline.

\section{A Detailed Expert Knowledge Description for the Text-guided Semantic Decoupling Module}

When using the text-guided semantic decoupling module, we encode multiple expert knowledge descriptions corresponding to each disease and compute the average of their embedding vectors as the final disease-wise text embedding for that disease. Table \ref{tab:expert describe} lists multiple expert knowledge descriptions corresponding to each disease, which are directly borrowed from \cite{SilvaRodriguez2025}.




\end{document}